# End-to-End Topology-Aware Machine Learning for Power System Reliability Assessment

Yongli Zhu, *Member, IEEE*, Chanan Singh, *Fellow, IEEE*

Texas A&M University, College Station, USA (yongliz@tamu.edu, singh@ece.tamu.edu)

*Abstract*—Conventional power system reliability suffers from the long run time of Monte Carlo simulation and the dimension-curse of analytic enumeration methods. This paper proposes a preliminary investigation on end-to-end machine learning for directly predicting the reliability index, e.g., the Loss of Load Probability (LOLP). By encoding the system admittance matrix into the input feature, the proposed machine learning pipeline can consider the impact of specific topology changes due to regular maintenances of transmission lines. Two models (Support Vector Machine and Boosting Trees) are trained and compared. Details regarding the training data creation and preprocessing are also discussed. Finally, experiments are conducted on the IEEE RTS-79 system. Results demonstrate the applicability of the proposed end-to-end machine learning pipeline in reliability assessment.

*Keywords*— reliability assessment, topology change, support vector machine, boosting trees, Monte Carlo simulation

## I. INTRODUCTION

Due to the increasing load demands and economic development, modern transmission power systems are subjected to various risks. Thus, the guarantee of continuously load supply-ability plays a vital role. To that end, power system reliability assessment is needed, which typically includes resource adequacy assessment (e.g., generation adequacy assessment) and system security assessment (e.g., N–1 analysis). This paper focuses on the generation adequacy assessment of bulk power systems (i.e., considering transmission networks). Conventional reliability assessment methods mainly include analytic enumeration, Monte Carlo simulation (MCS), and hybrid methods that combine the merits of the above two [1]. Analytical enumeration suffers from the exponentially growing enumeration space when the power grid becomes larger. In contrast, Monte Carlo simulation is theoretically dimension-free of the total number of system states. It typically induces three stages: a) sampling of the "state" (in non-sequential MCS) or "event" (in sequential MCS); b) testing of the state (i.e., to tell whether the state is "reliable/adequate" or not; c) updating statistics (i.e., interested reliability indices). The hybrid method combines the previous two methods, e.g., first applying state-space reduction strategies, then conducting MCS [2].

The MCS method has been de facto the standard practice for reliability assessment in the power industry for multi-area and composite systems since the 1990s [3]. However, all MCS methods suffer the time cost issues: when stricter convergence criteria are adopted, it takes more simulation runs (or called *replicas*). The bottlenecks are mainly in the state-sampling stage and state-testing stage. Thus, specific *variance reduction* techniques from the statistics area were employed to speed up the MCS. For example, importance sampling, stratified sampling, and dagger sampling [4].

On the other side, researchers have made efforts in applying machine learning (ML) methods for reliability assessment. In [5], the author uses transfer learning techniques to adapt the learned binary classifier for varying system load and installed capacities. In [6], a multi-label classifier is trained to simultaneously output the reliability indices for all the buses. In [7], the Genetic Algorithm (GA) has been successfully applied for: a) intelligent state selection (i.e., select more vulnerable states); b) binary encoded GA is utilized as a state sampling tool and c) replacing the DCOPF (direct current optimal power flow)-based load curtailment step by a real value-encoded GA model. In [8], a dynamic Bayesian belief network model is embedded into a Monte Carlo simulation-based unit commitment framework to predict the solar and wind generation for reliability evaluation.

The above methods focus on the state-testing stage and stage-sampling stage. However, they still need to be embedded in an MCS framework. In other words, those meta-heuristic and machine learning methods are not end-to-end: a large number of simulations or samplings are still required in the succeeding stage to obtain the final reliability indices. Thus, end-to-end machine learning can be considered. For example, in [9], the natural language processing method is applied to predict the distribution system's outage duration. In [10], the lower-level power converter reliability metrics are mapped to the overall system reliability metric by machine learning regression models.

This paper proposes an end-to-end ML-based pipeline for power system reliability assessment. The proposed pipeline can tackle the challenge of topology change in the studied power grid. The proposed method can save considerable time compared to regular Monte Carlo methods. Section II briefly introduces the basics of MCS for power system reliability assessment. Section III and IV explain the proposed end-to-end, topology-aware ML pipeline in detail. Section V presents the results of different experiments on the IEEE RTS-79 system. Conclusions and next steps are given in Section VI.

## II. MONTE CARLO SIMULATION FOR POWER SYSTEM RELIABILITY ASSESSMENT

### A. Non-sequential Monte Carlo Simulation

In power system reliability assessment, Monte Carlo simulation includes non-sequential and sequential types based on different simulation philosophies. In this paper, the non-sequential MCS is utilized. The basic MCS procedure for a system with *m* components is described as follows:

Stage-1: Sampling a system state $s = (s_1, …, s_m)$ based on each component's failure rate or probability distribution;

Stage-2: Testing the state by certain criteria such as DCOPF or other simple rules (e.g., comparing the total generator capacities with the total load);

Stage-3: Update the reliability index and its variance:

$$E[F] = \frac{1}{N}\sum_{j=1}^{N} F(s^j), \quad Var[F] = \frac{1}{N}\sum_{j=1}^{N} (F(s^j) - E[F])^2 \quad (1)$$

where $s^j$ is the system state sampled at the $j$-th time; $N$ is the total simulation runs; $E[\cdot]$ means the expected value. $F$ is the mapping function related to a specific reliability index. For example, if Eq. (2) is adopted for the function $F$, then the expected value in Eq. (1) will be the LOLP (Loss of Load Probability).

$$F(s^j) = \begin{cases} 0 & s^j \text{ is reliable state} \\ 1 & s^j \text{ is a failure state} \end{cases} \quad (2)$$

*B. Testing of States in MCS by ACOPF*

When a system state is obtained after a failure event of generators is sampled, the next step is to test the state by certain modified optimal power flow since the transmission network is now considered [1]. Compared to the DCOPF, more comprehensive requirements (e.g., the limit for the bus voltage magnitude) can be considered by the ACOPF (alternative current optimal power flow). Thus, this paper adopts a *tailored* ACOPF, where the constraints are the same as that in a regular ACOPF (e.g., voltage magnitude, generator var limit, and the branch overflow constraints). But the objective function is preset to be zero.

Here, the tailored ACOPF actually plays the role of system operators to mimic the mitigation actions when a generator failure event happens. If this tailored ACOPF finally diverges, a "failure state" is indicated for the tested state; otherwise, a "reliable state" is indicated.

*C. Stop-Criteria*

As shown in Eq. (3), the "coefficient-of-variation" (denoted as "$\beta$") is typically adopted as the stop-criteria in the (non-sequential) MCS procedure for reliability assessment [1]. When $\beta$ is less than a certain threshold (e.g., 0.01 to 0.05), the MCS procedure will terminate, and the result in Eq. (1) will be output as the final reliability index.

$$\beta \triangleq \frac{\sqrt{Var[E[F]]}}{E[F]}, \quad Var[E[F]] = \frac{1}{N} Var[F] \quad (3)$$

However, a too-small threshold value for $\beta$ typically results in an unnecessarily long running time of the MCS procedure, increasing the overhead cost of obtaining the training dataset for the machine learning purpose. On the other hand, a too large $\beta$ threshold can lead to inaccurate reliability index estimation. In reality, a maximum iteration number and a reasonably small $\beta$ threshold can be combinedly used in an "or" condition.

## III. END-TO-END MACHINE LEARNING PIPELINE FOR RELIABILITY ASSESSMENT

*A. Basic Idea*

As discussed in Section I, previous research efforts are mainly associated with the sampling and testing stages. Such ideas try to mimic the OPF behavior: the output of the trained machine learning model is either the optimal load curtailment or simply a binary indicator regarding the sampled system state (i.e., reliable or not). Those model outputs will then be utilized inside the regular MCS procedure during the inference stage. Thus, it can be named as "ML-embedded Monte Carlo" method. Its basic workflow is depicted in Fig. 1. The model input **X** is a specific mapping of system states or measurements (e.g., voltage, power, branch impedance), and the model output **Y** is a binary vector.

Compared to the conventional MCS method, this "indirect style" workflow can reduce the total time cost. However, there are certain limitations:

(1) MCS is still required in the outer loop. Thus, the overall time cost can still be large.

(2) The internal ML model suffers the notorious "class unbalanced" issues: in the reliability simulation, the number of *negative samples* (unreliable states with loss of load) is typically much *rarer* than that of the positive samples (reliable states). This limitation requires extra handling, such as down-sampling or other sophisticated machine learning algorithms.

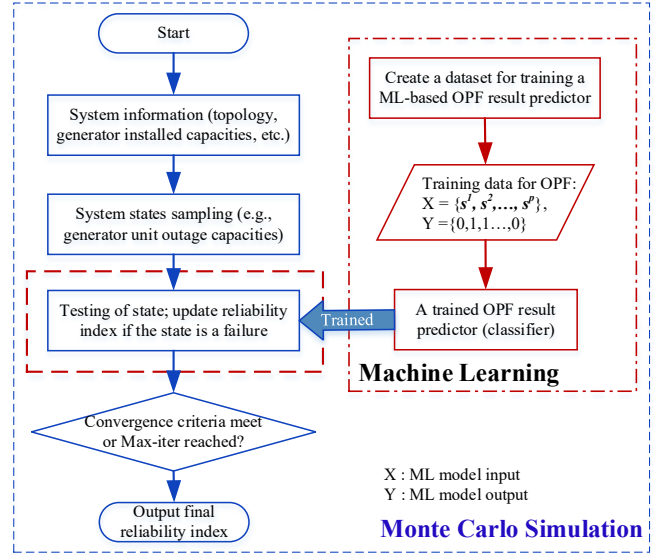

Fig. 1. ML-embedded Monte Carlo simulation for reliability assessment

Therefore, an end-to-end machine learning pipeline can be adopted to tackle the above challenges, directly predicting the desired reliability index values (e.g., LOLP). The proposed end-to-end machine learning pipeline is depicted in Fig. 2.

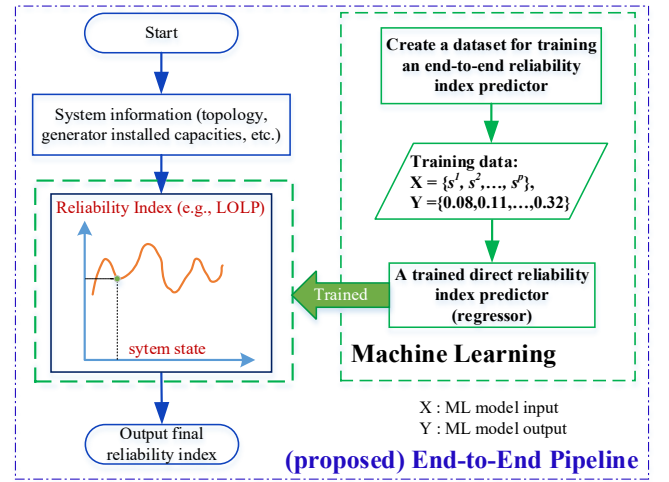

Fig. 2. End-to-End ML-based pipeline for reliability assessment

*B. Feature Engineering*

*1) Scheme-1: Raw features*

In machine learning, the selected features of the input data are typically expected to contain as much physical information as possible. However, not all the information is helpful during the reliability index calculation. For example, in reliability assessment, the installed capacity $P_{g,max}$ rather than the active power $P_g$ is the quantity needed for each generator unit.

The Y-bus matrix (its real part **G** and imagery part **B**) is utilized to consider the impact of network topology change (i.e., "topology-aware"). The matrix **G** is considered here since ACOPF rather than DCOPF is adopted for dataset creation. Initially, each input data sample can have the following arrangement:

$$X = [\mathbf{G}, \mathbf{B}, P_d, Q_d, P_{g,max}] \quad (4)$$

where **G** and **B** are *n*-by-*n* matrices; $P_d$, $Q_d$ are *n*-by-1 vectors for each load bus's active and reactive power; $P_{g,max}$ is an *n*-by-1 vector with non-generator bus positions filled with zeros. Thus, the final shape of each input data sample is *n*-by-(2*n*+3). However, scheme-1 has the following drawbacks:

- Wasting of memory space due to the symmetric property of the matrices **G** and **B**;
- Slow training for the machine learning model due to those extra input parameters.

Thus, it will *not* be considered in this paper.

*2) Scheme-2: Full features*
- Extracting the upper triangular part of **G** as a 1-dim column vector $v_G$. Similarly, obtaining $v_B$ from **B**.
- Concatenating all the 1-dim vectors together.

Finally, nearly half of the original space can be saved, and the new shape of each data sample is $n(n+4)$-by-1, as shown in Eq. (5). In this way, it is named "full features".

$$X = [v_G; v_B; P_d; Q_d; P_{g,max}]^T \quad (5)$$

*3) Scheme-3: Partial features*

To further reduce the input feature size, Eq. (6) is adopted, i.e., discard all the non-diagonal elements of the two matrices since most of those elements are zeros in the original matrices. In this way, it is named "partial features".

$$X = [\mathrm{diag}(\mathbf{G}); \mathrm{diag}(\mathbf{B}); P_d; Q_d; P_{g,max}] \quad (6)$$

The above two feature engineering schemes are illustrated in Fig. 3.

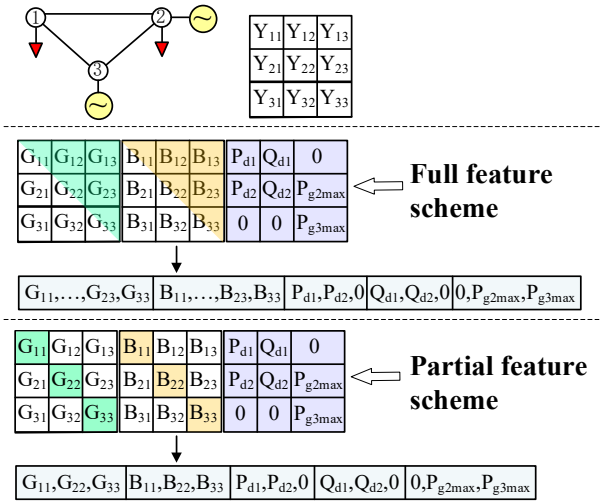

Fig. 3. Illustrations of two feature engineering schemes considering system topology information

## C. Dataset Creation

The regular Monte Carlo method is utilized to create the dataset for adequacy evaluation. To incorporate the impact of system topology change during dataset creation, MCS is conducted for all cases with first-order topology change (i.e., removal of any one branch) and second-order topology change (i.e., simultaneously removing two branches). Higher-order (≥3) topology changes are not considered since they are uncommon in the utility maintenance schedule for branches.

It should be clarified that the topology change considered here is set up *before* entering the MCS since it is adopted to mimic the planned outages of certain branches (line or transformer) for routine maintenance. It is *not* identical to the "branch unavailability" event in the internal loop of the MCS (if transmission line outages are considered). Besides, the following two steps are considered:

Step-1: Before entering the Monte Carlo simulation, check the grid connectedness [11] after applying the first- or second-order topology changes and excluding the disconnected cases;

Step-2: After Step-1 and *before* entering the Monte Carlo simulation, each case will be conducted a one-time ACOPF run. Any cases that fail this initial ACOPF test will be excluded from the succeeding Monte Carlo simulation and the final dataset (since those cases will always fail the ACOPF, thus, their LOLP values are always equal to 1).

After these two steps, the dataset finally becomes more *learnable* from a machine learning perspective.

## D. Feature Normalization

Here, "normalization" means scaling the feature to a fixed range. By trial-and-error, [0,1] scaling is applied to input data by Eq. (7), where "*x*" represents a feature vector or matrix.

$$x = (x - x_{\min}) / (x_{\max} - x_{\min}) \quad (7)$$

Because the selected feature variables have different physical meanings and ranges, thus they are scaled separately to avoid numerical issues, i.e., **G**, **B**, $P_d$, $Q_d$, and $P_{g,max}$ are scaled using their *max* and *min* elements, respectively.

## IV. MACHINE LEARNING MODELS

This paper considers two ML models for regression: Support Vector Machine (SVM) and Boosting Trees (BT). Both the two models are implemented in MATLAB. It is worth mentioning that, if necessary, more efficient ML tools like Python *sklearn* can be used to achieve potentially better accuracy or speed.

## A. Support Vector Machine (SVM)

SVM is an ML model with good interpretability and a reasonable amount of hyperparameters. Most of all, it can offer good generalizability for a small amount of data [12]. Its primal formulation for regression is shown in Eq. (8):

$$\begin{aligned}
\min J(w, b, \xi_i, \xi_i^*) &= \frac{1}{2}\|w\|_2 + C\sum_{i=1}^{N}(\xi_i + \xi_i^*) \\
s.t.\quad y_i - f(x_i) &\leq \varepsilon + \xi_i, \quad i = 1..N \\
f(x_i) - y_i &\leq \varepsilon + \xi_i^*, \quad i = 1..N \\
0 \leq \xi_i,\ 0 &\leq \xi_i^* \quad\quad i = 1..N \\
f(x) &= \langle w, x \rangle + b
\end{aligned} \quad (8)$$

where: *N* is the total number of data samples. $x_i$ is the *i*-th data sample, and $y_i$ is the actual response value for the *i*-th data sample. *w* and *b* are the parameters to be learned. *C* is a positive *regularization* constant controlling the penalty imposed on samples outside the margin (*ε*). A suitably large *C* value can prevent overfitting. $\xi_i$ and $\xi_i^*$ are slack variables introduced for each sample. More details can be found in [12].

### B. Boosting Trees (BT)

The Boosting Trees (BT) model combines many decision trees to reduce the risk of overfitting in every single tree. It utilizes an idea called *boosting* [12], i.e., connecting *weak learners* (usually decision trees with only one split) sequentially such that each new tree corrects the errors of the previous one. One representative version of the BT model, *Least Square Boost*, is shown as follows.

| Algorithm: Least Square Boost |
|---|
| 1 **Input**: $(\mathbf{x}_i, \mathbf{y}_i)$, $i = 1…N$ |
| 2 $F_0(\mathrm{x}) = \bar{y}$    // using average value for initialization |
| 3 **for** *m =1 to M*: |
| 4    $\tilde{y}_i \leftarrow \bar{y}_i - F_{m-1}(\mathbf{x}_i)$, $i = 1…N$ |
| 5    $(\rho_m, \boldsymbol{\theta}_m) \leftarrow \mathrm{argmin}_{\rho,\theta} \sum_{i=1}^{N}[\tilde{y}_i - \rho h(\mathbf{x}_i; \boldsymbol{\theta})]^2$ |
| 6    $F_m(\mathbf{x}) \leftarrow F_{m-1}(\mathbf{x}) + \rho_m h(\mathbf{x}; \boldsymbol{\theta}_m)$ |
| 8 **Output** $F_M(\cdot)$ as the final regressor |

In the above algorithm: $(\mathbf{x}_i, \mathbf{y}_i)$ ($i = 0$ to $M$) are the training dataset. $F_i$ ($i = 0$ to $M$) is a series of tree regressors, and each of them is trained based on the residual of its predecessor. *h* is the bottom level regressor (formally parametrized by **θ**), e.g., a simple decision tree. $\rho_m$ and $\boldsymbol{\theta}_m$ are the parameters to learn.

## V. CASE STUDY

### A. System Description

The proposed pipeline is tested on the IEEE RTS-79 system, as shown in Fig. 4. The original RTS-79 system has 24-bus, 32 generator units, and 38 branches.

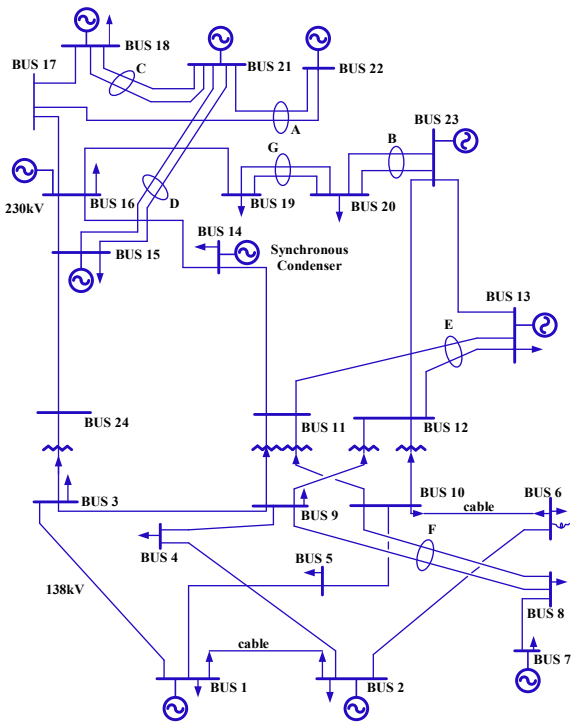

Fig. 4.   The single-line diagram of the IEEE RTS-79 system

All the generators are two-state probabilistic failure models based on the given "FOR" (forced-outage-rate). The *unavailability events* of generator units are considered up to three orders (i.e., simultaneously loss of three units) since the higher-order event typically has a lower chance in real-life and Monte Carlo simulations. Branches (transmission lines/transformers) in this study are considered available during the whole run since failure rates of those components are much lower than generators. However, the capacity constraints of transmission lines are still considered [6]. It should be clarified that 1) one *complete* MCS is run for *each* "data sample," and the only thing that is varied between each MCS's internal iteration is the randomly drawn generator outages; 2) since the impact of the topology change is the main concern in this paper, the loading variation is not considered.

Based on steps listed in Section III.C, the total number of finally considered 1st-order topology-change cases are 34 after excluding the divergent cases (in the initial ACOPF test) and the bridge cases. Likewise, the total number of finally considered 2nd-order topology-change cases is 343. In total, 343+34=377 samples are collected using about 21 hours running time of MCS (stop-criteria: *β*<0.02, maximum iterations=4000). Thus, the average single MCS-run costs nearly 200 sec. The dataset generation, ML model training, and performance testing are conducted on a computer with Intel CPU 6700K and 32GB memory. Note that the obtained LOLP is for the whole system rather than per bus.

### B. Basic Performance

To test the basic performance of the proposed end-to-end reliability predictor, a set of 10% of the data is randomly picked for testing, and the remaining 90% is left for training. Regarding performance validation and model comparison, three error metrics, i.e., MAPE (mean-absolute-percentage-error), MAE (mean-absolute-error), and RMSE (root-mean-square-error), are adopted in the following experiments.

By trial-and-error, the *linear kernel* function is selected to use in the SVM; the number of learners (*learning cycles*) is set to 100 for BT. The error metrics are shown in Table I. The LOLPs of the base case and four testing samples are shown in Table II as examples. From the results, SVM is superior to BT.

TABLE I.   BASIC PERFORMANCES OF SVM AND BT (BASELINE: MCS)

|  | MAPE | MAE | RMSE |
|---|---|---|---|
| **SVM** | 5.02% | 0.0047 | 0.0071 |
| **BT** | 8.69% | 0.0079 | 0.0138 |

TABLE II.   EXAMPLE LOLP COMPARISONS

|  | MCS | SVM | BT |
|---|---|---|---|
| Base case | 0.08467 | 0.08531 | 0.08955 |
| Sample 1 | 0.08800 | 0.08866 | 0.09123 |
| Sample 2 | 0.08800 | 0.08812 | 0.08695 |
| Sample 3 | 0.08975 | 0.08973 | 0.08293 |
| Sample 4 | 0.09275 | 0.09220 | 0.10643 |

### C. Scalability on Reduced Training Dataset

When the training data is insufficient, the performance of the pipeline may decline. To investigate this scalability concern, three different sizes of training datasets are examined, i.e., 90%, 70%, 50%, and the respectively remaining data are used for testing. The error metrics are shown in Table III. Comparisons for the base case LOLP are shown in Table IV. As expected, the performance of the ML model can deteriorate when training data size decreases; but on this specific problem, SVM still performs better than BT.

TABLE III. SCALABILITY ON REDUCED TRAINING DATA

| SVM | MAPE | MAE | RMSE |
|---|---|---|---|
| 90% | 5.02% | 0.0047 | 0.0071 |
| 70% | 6.11% | 0.0087 | 0.0350 |
| 50% | 7.24% | 0.0096 | 0.0345 |
| **BT** | MAPE | MAE | RMSE |
| 90% | 8.69% | 0.0079 | 0.0138 |
| 70% | 8.41% | 0.0105 | 0.0334 |
| 50% | 9.74% | 0.0129 | 0.0376 |

TABLE IV. BASE CASE LOLP COMPARISONS

| Training data | Base case LOLP | | |
|---|---|---|---|
| | SVM | BT | MCS |
| 90% | 0.08531 | 0.08955 | |
| 70% | 0.08645 | 0.08755 | 0.08467 |
| 50% | 0.08277 | 0.08823 | |

### D. Generalizability on Unseen Higher-Order Topology Changes

To investigate the generalizability of the proposed ML pipeline in situations of unseen higher-order topology changes, a set of 20 three-order line-removal samples are randomly generated for testing (recall that the original dataset contains only first- and second-order line-removal samples).

Again, three different sizes of the *original* training datasets are examined, i.e., 100%, 75%, and 50%. The error metrics are shown in Table V. Comparisons for LOLP of five three-order-topology-change samples are shown in Table VI as examples (100% original data is used for training).

It is not surprising that the MAPE of both methods becomes larger since all the testing samples are not only unseen during their "learning process" but also *heterogeneous*. The error metrics increase slightly for SVM but drastically for BT, especially on the small training dataset. Regarding the LOLPs, except for the 3rd sample, SVM results are closer to MCS results (baseline) than that of BT.

In a word, the results here imply that the proposed end-to-end ML pipeline still has certain forecasting power even in situations with heterogeneous topology changes.

TABLE V. GENERALIZABILITY ON UNSEEN THREE-ORDER SAMPLES

| SVM | MAPE | MAE | RMSE |
|---|---|---|---|
| 100% | 6.51% | 0.0062 | 0.0089 |
| 75% | 6.13% | 0.0059 | 0.0088 |
| 50% | 7.76% | 0.0072 | 0.0096 |
| **BT** | MAPE | MAE | RMSE |
| 100% | 8.36% | 0.0078 | 0.0123 |
| 75% | 16.53% | 0.0155 | 0.0254 |
| 50% | 14.90% | 0.0134 | 0.0241 |

TABLE VI. EXAMPLE LOLP COMPARISONS ON UNSEEN THREE-ORDER SAMPLES (100% ORIGINAL DATA FOR TRAINING)

| | MCS | SVM | BT |
|---|---|---|---|
| 3-order topo. change Sample 1 | 0.09050 | 0.08965 | 0.08718 |
| 3-order topo. change Sample 2 | 0.09025 | 0.08940 | 0.08295 |
| 3-order topo. change Sample 3 | 0.09625 | 0.09440 | 0.09693 |
| 3-order topo. change Sample 4 | 0.08300 | 0.08350 | 0.08429 |
| 3-order topo. change Sample 5 | 0.09100 | 0.08974 | 0.08955 |

### E. Effect of Using Partial Features

Here, the "partial features" scheme in Section III is used, and experiments in the last two subsections are redone. In Table VII and Table IX, the error metrics are close to that in the previous case where the "full features" scheme is adopted.

SVM still performs better than BT. In Table VIII and Table X, the base case LOLPs of SVM are, in general, better than that of BT except on few samples.

TABLE VII. SCALABILITY ON REDUCED TRAINING DATA SET (PARTIAL FEATURES)

| SVM | MAPE | MAE | RMSE |
|---|---|---|---|
| 90% | 5.06% | 0.0047 | 0.0072 |
| 70% | 6.18% | 0.0086 | 0.0339 |
| 50% | 6.47% | 0.0092 | 0.0349 |
| **BT** | MAPE | MAE | RMSE |
| 90% | 7.68% | 0.0069 | 0.0144 |
| 70% | 8.77% | 0.0101 | 0.0273 |
| 50% | 8.97% | 0.0110 | 0.0329 |

TABLE VIII. BASE CASE LOLP COMPARISONS (PARTIAL FEATURES)

| Training data | Base case LOLP | | |
|---|---|---|---|
| | SVM | BT | MCS |
| 90% | 0.08750 | 0.08864 | |
| 70% | 0.08778 | 0.08729 | 0.08467 |
| 50% | 0.08794 | 0.09361 | |

TABLE IX. GENERALIZABILITY ON UNSEEN THREE-ORDER SAMPLES (PARTIAL FEATURES)

| SVM | MAPE | MAE | RMSE |
|---|---|---|---|
| 100% | 6.71% | 0.0065 | 0.0091 |
| 75% | 6.62% | 0.0064 | 0.0092 |
| 50% | 6.89% | 0.0066 | 0.0092 |
| **BT** | MAPE | MAE | RMSE |
| 100% | 9.75% | 0.0091 | 0.0129 |
| 75% | 21.09% | 0.0199 | 0.0401 |
| 50% | 27.05% | 0.0244 | 0.0392 |

TABLE X. EXAMPLE LOLP COMPARISONS ON UNSEEN THREE-ORDER SAMPLES (100% ORIGINAL DATA FOR TRAINING) (PARTIAL FEATURES)

| | MCS | SVM | BT |
|---|---|---|---|
| 3-order topo. change Sample 1 | 0.09050 | 0.08847 | 0.08729 |
| 3-order topo. change Sample 2 | 0.09025 | 0.08731 | 0.08516 |
| 3-order topo. change Sample 3 | 0.09625 | 0.09671 | 0.09003 |
| 3-order topo. change Sample 4 | 0.08300 | 0.08280 | 0.08448 |
| 3-order topo. change Sample 5 | 0.09100 | 0.09217 | 0.09295 |

### F. Time Cost Comparison

The time costs of the two methods for the experiments in Table III and VII are depicted in Fig. 5. It is observed that SVM has demonstrated an overall speed advantage and performs even faster when using partial features.

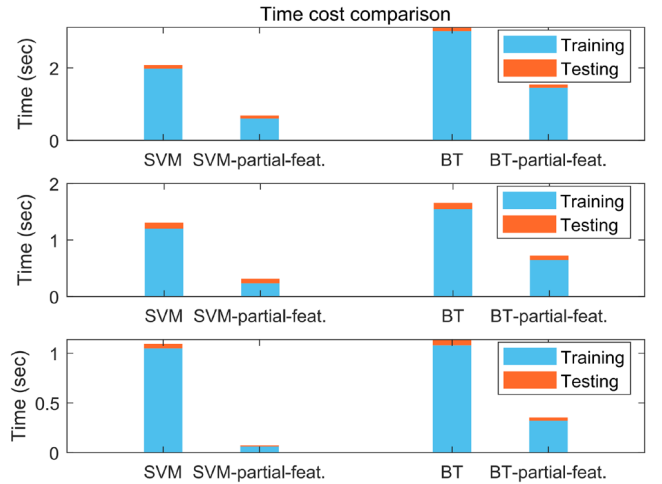

Fig. 5. Time comparison (SVM .vs. BT; full features v.s. partial features) (top, midle, bottom: 90%, 70%, 50% data for training)

As mentioned in subsection-A, one regular MCS-run takes about 200 sec for this system. Thus, "10% of testing data" will lead to a time cost of more than 2 hours (377*0.1*200/3600≈2.09); "30% of testing data" will lead to a time cost of more than 6 hours (377*0.3*200/3600≈6.28); "50% of testing data" will lead to a time cost more than 10 hours (377/2*200/3600≈10.47). Therefore, the proposed end-to-end ML pipeline can save considerable time compared to the conventional Monte Carlo method (even the training time cost is included).

## VI. CONCLUSION

The proposed end-to-end, topology-aware, machine learning pipeline demonstrates good scalability and generalizability in partial data, partial features, and unseen topology-change situations when using SVM. Its speed advantage over the regular Monte Carlo method is also verified. The next steps are: 1) fine-tuning model hyperparameters and investigating other machine or deep learning models for further performance improvement; 2) considering *transfer learning* to improve the pipeline's adaptability to other kinds of unseen topologies (e.g., adding a new transmission line).